# Complementary Pseudo Multimodal Feature for Point Cloud Anomaly Detection

Yunkang Cao, Xiaohao Xu, Weiming Shen

*Abstract*—Point cloud (PCD) anomaly detection steadily emerges as a promising research area. This study aims to improve PCD anomaly detection performance by combining handcrafted PCD descriptions with powerful pre-trained 2D neural networks. To this end, this study proposes Complementary Pseudo Multimodal Feature (CPMF) that incorporates local geometrical information in 3D modality using handcrafted PCD descriptors and global semantic information in the generated pseudo 2D modality using pre-trained 2D neural networks. For global semantics extraction, CPMF projects the origin PCD into a pseudo 2D modality containing multi-view images. These images are delivered to pre-trained 2D neural networks for informative 2D modality feature extraction. The 3D and 2D modality features are aggregated to obtain the CPMF for PCD anomaly detection. Extensive experiments demonstrate the complementary capacity between 2D and 3D modality features and the effectiveness of CPMF, with 95.15% image-level AU-ROC and 92.93% pixel-level PRO on the MVTec3D benchmark. Code is available on https://github.com/caoyunkang/CPMF.

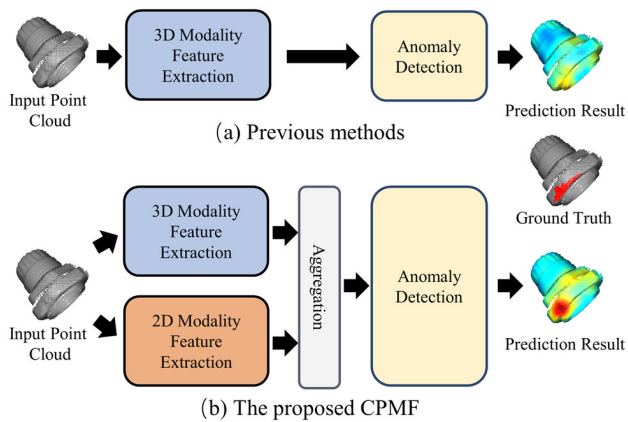

Fig. 1. Different from previous methods that only considers local geometric cues of 3D modality, the proposed method (CPMF) considers both local geometry of 3D modality and global semantics of 2D modality.

## I. Introduction

WITH the booming development of 3D collection sensors and devices, point cloud (PCD), which explicitly captures geometrical information of real-world objects, has drawn great popularity. Meanwhile, PCD anomaly detection, *i.e.*, detecting abnormal points that deviate from normal patterns for 3D structures, becomes a promising direction in industrial inspection [1]–[3] and medical diagnosis [4]. However, as the practice of using PCD for anomaly detection is just emerging, their performances are far from satisfactory for real-world applications. This study focuses on PCD anomaly detection and tries to meet practical performance requirements by collaboratively unifying local geometric and global semantic cues for PCD.

Previous anomaly detection methods for images [5] and graphs have shown that distinctive feature representations [6] are crucial for success. Similarly, recent PCD anomaly detection methods [1]–[3] have also recognized the importance of feature learning. Early PCD representations were handcrafted and relied on heuristic designs [7], [8]. With the development of deep learning, recent methods have adopted learning-based PCD features. Built on these prosperous PCD representations, 3D-ST [2] designs a self-supervised learning scheme for PCD anomaly detection. Despite considerable improvements versus baselines [1], its performance is still subpar. BTF [3] reveals the importance of feature descriptiveness in PCD anomaly detection. Contrary to expectations, the results show that classical handcrafted PCD descriptors outperform learning-based pre-trained features. BTF attributes this phenomenon to the domain distribution gap between target objects and pre-trained datasets, which results in a low transfer capability of the pre-trained PCD features. However, although classical handcrafted features achieve impressive performances, they are limited to using local structural information and are not able to access the global semantic context. As global semantic context is essential in detecting semantic abnormalities [5], [9], incorporating it with the geometrical modeling ability of handcrafted features can potentially bring improvements.

Compared to PCD anomaly detection, image anomaly detection [10], [11] is more sophisticated and has been studied more extensively. Early image anomaly detection methods [12] focus on learning descriptive features with only normal data and gradually develop excellent but unsatisfactory performance.

With a thorough investigation, recent methods [5], [13]–[16] find that representations from pre-trained 2D neural networks are adequate for anomaly detection, and they attribute this solid performance to the geometrical and semantic information hiding in pre-trained representations. Concretely, pre-trained image representations employ knowledgeable prior information from large-scale image datasets like ImageNet [17], so they are adept at capturing meaningful global semantic cues with impressive transfer capability.

Inspired by the remarkable attainment achieved by existing PCD and image anomaly detection methods, this study proposes a unified PCD representation, *i.e.*, Complementary Pseudo Multimodal Feature (CPMF), to fully exploit both the local geometric structure and global semantic context in PCD. Built upon handcrafted PCD descriptors, the proposed CPMF further exploits descriptive pre-trained 2D neural networks in a generated pseudo 2D modality to enrich the semantic content of the PCD descriptors. While existing methods [1]–[3] only work on the 3D modality for feature extraction, the proposed method captures complementary structural and semantic cues with the generated pseudo multimodal data and boosts the PCD anomaly detection performance, as shown in Fig. 1.

Technically, for 3D modality feature extraction, CPMF leverages classical handcrafted descriptors to capture precise local geometrical information in PCD. A 2D modality feature extraction module, which employs the advanced describing capability of pre-trained 2D neural networks, is further incorporated by CPMF to exploit and complement semantic information hidden in PCD. Particularly, this module utilizes 3D to 2D projection and rendering to translate the origin 3D PCD into the pseudo 2D modality containing multi-view 2D images. Then this module utilizes pre-trained 2D neural networks to extract strong semantic feature maps of these multi-view images. Subsequently, in the 2D to 3D alignment and aggregation process, the feature maps are mapped into point-wise features according to the 2D-3D projection correspondence in individual views. Those single-view point-wise features are further aggregated to obtain final 2D modality point-wise features composing cues from different views. The extracted 3D and 2D modality features are highly complementary because 1) The 3D handcrafted features are good at describing local structures but cannot exploit global semantic information [3], [7]. 2) The 2D pre-trained networks contain profound knowledge from large-scale image datasets and can capture semantic properties but fail to represent local structures precisely [5], [15]. Hence, CPMF develops an aggregation module to fuse 3D and 2D modality features and obtains features containing both global semantic and local geometrical information. Fig. 1 shows that CPMF performs better qualitatively than prior methods [3].

The contributions of this study are summarized as follows:
- This study addresses the problem of point cloud (PCD) anomaly detection and develops a Complementary Pseudo Multimodal Feature (CPMF) method, collaboratively unifying local geometric and global semantic cues in PCD by extracting PCD features in the pseudo multi-modality.
- This study proposes a novel way to obtain point-wise semantic PCD features by using pre-trained 2D networks in a pseudo 2D modality generated through projection and rendering. The extracted 2D modality features prove excellent PCD describing capacity.
- This study proves the complementation capacity between 3D and 2D modality features and demonstrates that CPMF achieves excellent PCD anomaly detection performance, outperforming existing PCD anomaly detection methods by a large margin.

The structure of this paper is organized as follows. The related work is comprehensively reviewed in Section II. Section III elaborates on the proposed CPMF. Experiments are illustrated and analyzed in Section IV. Finally, Section V concludes this paper and discusses future research directions.

## II. RELATED WORK

The related work is reviewed under three topics: 2D image anomaly detection, 3D PCD anomaly detection, and PCD feature learning.

### A. 2D Image Anomaly Detection

2D image anomaly detection is a widely-investigated task, especially after establishing the industrial inspection dataset MVTec AD [10]. 2D image anomaly detection methods typically consist of two modules, *i.e.*, feature extraction and feature modeling. The feature extraction module strives for features with better descriptiveness; ideally, the normal and abnormal features distribute differently and are easily distinguished. The feature modeling module is trained to model the distribution of normal features and is expected to deviate when inputting abnormal features.

Early work [12], [18]–[21] learned descriptive features from scratch. AESSIM [19] utilized an auto-encoder and an SSIM loss function with a reconstruction-based proxy task to learn features. RIAD [12] further developed an inpainting task to learn better representations. Then, DRAEM [18] developed a denoising reconstruction strategy and directly predicted the segmentation map for abnormalities. Differently, Cutpaste [20] built a self-supervised feature learning scheme via abnormality synthesis.

Despite the considerable performance of early methods, representations from pre-trained networks are proven to be effective and fascinating for anomaly detection [5], [13]–[16], [22], [23]. ST [15], IKD [5], and AST [16] all utilized knowledge distillation for normal feature distribution modeling. ST [15] aligned the normal features between a pre-trained teacher network and a randomly initialized student network. The alignment errors were utilized to score anomalies. IKD [5] achieved better describing capability with a context similarity loss function and a dynamic hard sample mining strategy. AST [16] alleviated the overgeneralization problem for anomaly detection by using an asymmetric teacher-student network architecture. Similarly, CDO [22] generated synthetic abnormalities and developed a novel collaborative discrepancy optimization loss function to address overgeneralization. Other methods investigated more potentials, like normalizing flow [22] and memory banks, for normal feature distribution modeling.

## B. 3D PCD Anomaly Detection

Notwithstanding profound studies on 2D image anomaly detection, 3D PCD anomaly detection has not been thoroughly investigated. The popularity of 3D PCD anomaly detection was boosted after the publication of the first real-world 3D anomaly detection dataset, MVTec 3D [1].

The feature modeling modules for PCD anomaly detection are largely influenced by image anomaly detection methods. However, as the properties of image and PCD are highly different, the feature extraction module in PCD anomaly detection needs to be newly designed. Hence, recent PCD anomaly detection methods highlight the importance of feature learning. The baselines [1] conducted a comprehensive benchmarking for previous competitive methods, such as the variation model [24] and autoencoder [25]. 3D-ST [2] developed a self-supervised learning scheme for PCD representation learning and employed a knowledge distillation-based modeling module for anomaly detection. Moreover, BTF [3] emphasized the effectiveness of classical handcrafted PCD descriptors with comprehensive experiments but observed that learned features performed poorly. BTF attributed this counterinitiative situation to the insufficient transfer capacity of existing pre-trained features on existing small-scale PCD datasets.

Although handcrafted features achieve comparable anomaly detection performance, they can rarely capture semantic information in PCD data because of their heuristic designs. Since semantic information is crucial for anomaly detection, this study targets to complement the lack of semantics and generate better PCD representations.

## C. PCD Feature Learning

Extensive literature has investigated PCD representations for popular 3D applications such as 3D grasp detection [26] and 3D registration [8].

The majority of existing methods extract PCD features in 3D modality directly. Classical methods [7], [8] rely on heuristic designs to model geometrical information. For instance, FPFH [7] is a well-known robust and efficient 3D descriptor that incorporates point coordinates and surface normals to represent the properties of PCD. A more advanced electrostatic field theory-based point descriptor was proposed in [8] and deployed for freeform surface parts quality inspection. Later, many studies developed neural network architectures [27]–[29] for PCD representation learning. PointNet [27] and its extensions, such as PointNet++ [28], proposed to process PCD data and learn point-level representation in an end-to-end manner. Based on these backbones for PCDs, the community designs better feature learning methods such as self-supervised learning. FCGF [30] learned local feature descriptors through a fully-convolutional design and point-level metric learning. PointContrast [31] further improved FCGF and applied it to semantic understanding, developing better semantic representations.

By contrast, some work took advantage of well-developed 2D representations by rendering the original PCD into a 2D modality composing multi-view images. Multiview CNN [32] pioneered the investigation of PCD representations via 2D images. It transformed the PCD into multi-view images, which were inputted to Multiview CNN to extract instance-level features. Local shape descriptors [33] learned point-wise features via part correspondences. Concretely, it rendered local point patches into multi-view images and then applied metric learning to these images for patch-wise descriptive point feature learning. Following methods [34], [35] emphasized the importance of view selection and aggregation.

This study targets the PCD anomaly detection task, especially descriptive PCD feature learning. It proposes a method that unifies direct 3D modality and the generated pseudo 2D modality to represent PCD data comprehensively. Due to the complementation capability between 2D and 3D modality features, the final PCD representations incorporate local structural and global semantic knowledge and boost the performance of PCD anomaly detection.

## III. PROPOSED METHOD

In this section, the method of Complementary Pseudo Multimodal Feature (CPMF) is explained in detail. Firstly, the problem of PCD anomaly detection is defined. Then, the pipeline of CPMF is introduced. Subsequently, the key components of the proposed CPMF, including 3D modality feature extraction, 2D modality feature extraction, aggregation operation, and the corresponding anomaly detection process, are presented.

## A. Problem definition

Given a 3D PCD with $N$ points $\boldsymbol{P}_{3D} = \{p_1, p_2, ..., p_N \in \mathbb{R}^3\}$ as input, the goal of PCD anomaly detection is to output real-valued anomaly scores $\boldsymbol{S} = \{s_1, s_2, ..., s_N\}$ for all points, in which abnormal points are assigned with larger scores. Then, the anomalous regions in $\boldsymbol{P}_{3D}$ can be localized according to the anomaly scores. Typically, a PCD set $\mathbb{P}_{train} = \{\boldsymbol{P}_{3D}\}$ containing only normal PCDs is accessible in the training stage. In the testing stage, anomaly scores $\mathbb{S} = \{\boldsymbol{S}\}$ for the testing unknown PCD set $\mathbb{P}_{test} = \{\boldsymbol{P}_{3D}^{test}\}$ are required as outputs.

## B. Pipeline Overview

Existing PCD anomaly detection methods [2], [3], [16] focus on extracting high-distinctive PCD features. However, their extracted features are subpar and hardly enough for impressive anomaly detection. To fully exploit the underlying information and properties of PCDs, this study proposes a novel feature named CPMF, which employs both local geometric and global semantic representations via sophisticated handcrafted 3D PCD descriptors and pre-trained 2D neural networks, respectively.

As shown in Fig. 2, CPMF mainly consists of a 3D modality feature extraction module and a 2D modality feature extraction module. Both of which can extract point-wise descriptive PCD features. CPMF first delivers an input PCD $\boldsymbol{P}_{3D}$ to these two modules to extract 3D modality features $\boldsymbol{F}_{3D}$ and 2D modality features $\boldsymbol{F}_{2D}$, respectively. For 3D modality feature extraction, handcrafted PCD descriptors are directly utilized to represent local structures of $\boldsymbol{P}_{3D}$. The 2D modality feature extraction module first projects the original 3D PCD $\boldsymbol{P}_{3D}$ to the 2D modality containing multi-view images, which are further sent to pre-trained 2D neural networks to extract distinctive semantic feature maps. Then, these feature maps are aligned to 3D space to generate point-wise 2D modality features $\boldsymbol{F}_{2D}$. $\boldsymbol{F}_{3D}$

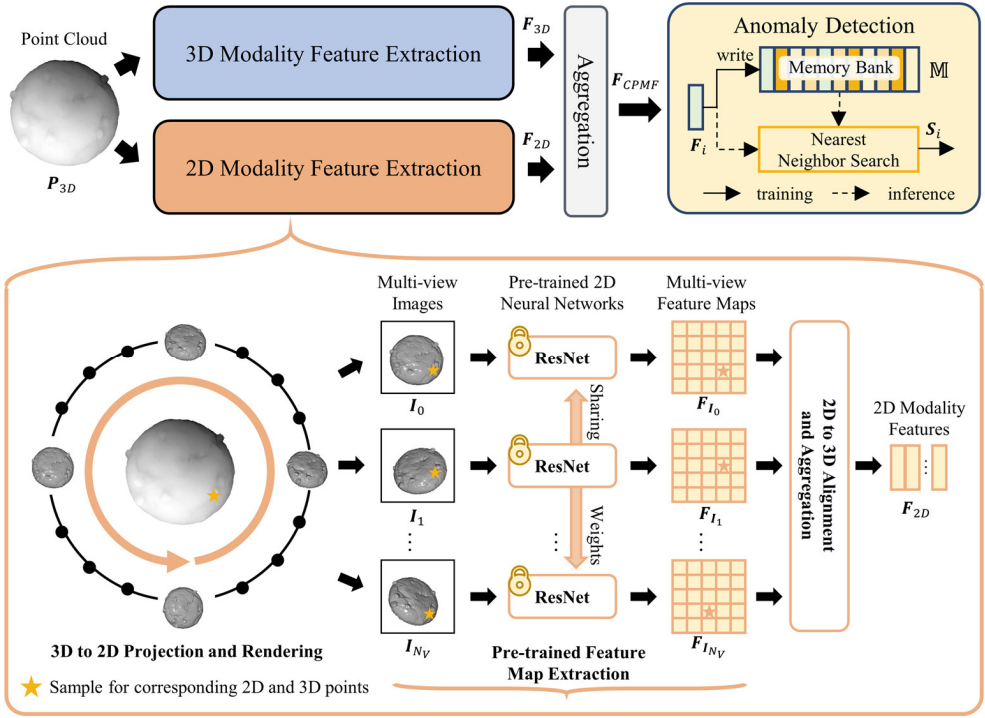

Fig. 2. The framework of the proposed CPMF. CPMF cooperates pseudo multimodal, i.e., 3D and 2D modalities, for PCD feature extractions. In 3D modality, CPMF leverages hand-crafted PCD descriptors for precise local geometrical representation. In 2D modality, CPMF renders the input PCD into multi-view images and fascinates the capability of pre-trained 2D neural networks for global semantic features extraction. Then complementary multimodal features are aggregated for final anomaly detection.

and $F_{2D}$ contain complementary knowledge as they exploit the information of $P_{3D}$ in different ways. Specifically, $F_{3D}$ considers local information between the neighboring points, and $F_{2D}$ contains strong semantic information inherited from pre-trained 2D neural networks with large receptive fields and profound semantic knowledge [15]. CPMF aggregates them together and generates complementary pseudo multimodal features $F_{CPMF}$. Finally, $F_{CPMF}$ cooperated with PatchCore [23], an excellent anomaly detection method using the nearest neighbor search, is leveraged for PCD anomaly detection.

In the next subsections, this paper will subsequently depict the 3D modality feature extraction module, the 2D modality feature extraction module, the aggregation operation, and the anomaly detection process.

### C. 3D Modality Feature Extractor

PCD feature extraction has been long investigated, from early handcrafted features [7], [8], [36] to recent learning-based deep features [2], [30]. Although learning-based features have presented impressive performances on some tasks, they are proven to be less effective for PCD anomaly detection than handcrafted features [2], [3], [7]. Hence, this study follows existing methods [3] and utilizes handcrafted descriptors to directly exploit the local geometrical information of PCD in 3D modality. Denoting the 3D feature extracting function as $\mathcal{F}_{3D}$, then 3D features $F_{3D}$ are obtained by,

$$F_{3D} = \mathcal{F}_{3D}(P_{3D}) \quad (1)$$

where $F_{3D} \in \mathbb{R}^{N \times D_{3D}}$, and $D_{3D}$ is the dimension of 3D modality features.

### D. 2D Modality Feature Extractor

Built upon handcrafted PCD descriptors, to further enhance the anomaly detection performance, this study proposes to generate pseudo 2D modality point-wise features via pre-trained 2D neural networks [5], [37], [38] and 2D-3D projection correspondences. Specifically, for an input PCD, the 3D to 2D projection and rendering operation is first employed to generate rendered multi-view images $\{I_1, I_2, ..., I_{N_V}\}, I \in \mathbb{R}^{H \times W \times 3}$ from different views, where $H \times W$ is the spatial resolution of rendered images, and $N_V$ is the number of rendering views. Then, for the pre-trained feature map extraction, pre-trained 2D neural networks are employed to extract feature maps for the multi-view images. For later 2D to 3D alignment, the feature maps are up-sampled to the same resolution as the original images, and the up-sampled feature maps are denoted as $\{F_I\}, F_I \in \mathbb{R}^{H \times W \times D_{2D}}$, where $D_{2D}$ is the dimension of the extracted features. Ultimately, image feature maps $\{F_I\}$ are aligned to the original 3D space and obtain point-wise 2D modality features $F_{2D} \in \mathbb{R}^{N \times D_{2D}}$.

#### 1) 3D to 2D Projection and Rendering

To take advantage of representations from well-developed pre-trained 2D neural networks, this study generates multi-view images with image rendering operation and feeds these images to 2D neural networks to extract multi-view image feature maps. Given a camera model parameter set $C$, the render function $\mathcal{R}$,

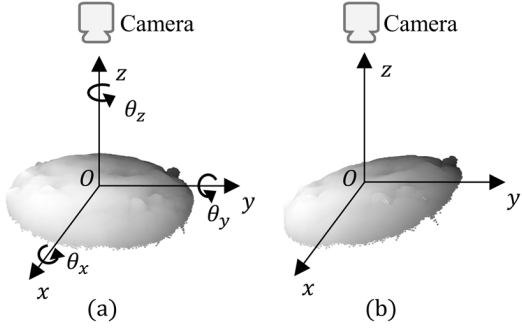

Fig. 3. (a) Illustration for PCD rotations of a cookie. (b) A sample of a rotated PCD of a cookie.

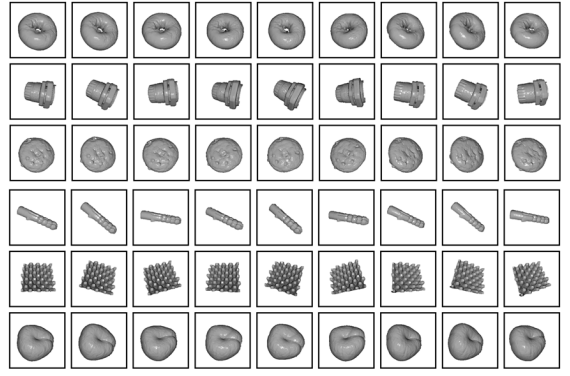

Fig. 4. Samples for multi-view rendered images (the first image is rendered in the original view).

which renders a PCD $P_{3D}$ into an image $I$, can be formulated as,

$$I = \mathcal{R}(P_{3D}, C) \quad (2)$$

For further 2D to 3D alignment and aggregation, it is necessary to acquire the correspondence between the spatial position of a 3D point and its location in 2D pixel coordinates, i.e., 2D-3D projection correspondences. Let $P_{3D,i}$ be the 3D position of the $i$-th point, and $P_{2D,i}$ be the corresponding locations in pixel coordinates, $P_{2D} \in \mathbb{R}^{N\times 2}$, the 2D-3D projection correspondences can be formulated as,

$$[P_{2D,i}, 1]^T = \frac{1}{Z_c} KT[P_{3D,i}, 1]^T \quad (3)$$

where $K$ and $T$ are the intrinsic and extrinsic matrixes of the rendered camera, respectively; $1/Z_c$ is the normalized coordinate coefficient.

As the rendering operation projects $P_{3D}$ into a lower dimension, i.e., from 3D to 2D, it may cause certain information losses [32]. To maintain intact information of $P_{3D}$, this study sequentially rotates $P_{3D}$ with different rotation matrixes. Typically, this study set various rotation matrixes $\{R_1, R_2, ..., R_{N_V}\}, R \in \mathbb{R}^{3\times 3}$ for image rendering. For the $i$-th point of the $k$-th rotated PCD $P_{3D,k,i}$ is,

$$P_{3D,k,i} = R_k P_{3D,i} \quad (4)$$

and its corresponding rendered image $I_k$ and 2D positions $P_{2D,k}$ are obtained by,

$$I_k = \mathcal{R}(P_{3D,k}, C) \quad (5)$$

$$[P_{2D,k,i}, 1]^T = \frac{1}{Z_c} KT[P_{3D,k,i}, 1]^T \quad (6)$$

Specifically, the rotation matrixes $R$ are obtained by selecting different angles $\theta_x, \theta_y, \theta_z$ for the $x,y,z$ axes, and $R$ is formulated by,

$$R = \begin{bmatrix} \cos\theta_x & \sin\theta_x & 0 \\ -\sin\theta_x & \cos\theta_x & 0 \\ 0 & 0 & 1 \end{bmatrix} \begin{bmatrix} \cos\theta_y & 0 & \sin\theta_y \\ 0 & 1 & 0 \\ -\sin\theta_y & 0 & \cos\theta_y \end{bmatrix}$$
$$\begin{bmatrix} 1 & 0 & 0 \\ 0 & \cos\theta_z & \sin\theta_z \\ 0 & -\sin\theta_z & \cos\theta_z \end{bmatrix} \quad (7)$$

Fig. 3 shows the details for PCD rotation and rendering. With the generated rotation matrixes $\{R_1, R_2, ..., R_{N_V}\}$, a sequence of rendered images $\{I_1, I_2, ..., I_{N_V}\}$ and their corresponding 2D locations $\{P_{2D,1}, P_{2D,2}, ..., P_{2D,N_V}\}$ are obtained using (4), (5), and (6). Fig. 4 shows some examples of the rendered images.

#### 2) Pre-trained Feature Map Extraction

With multi-view rendering, images from multiple views are acquired. Then, this study delivers them to pre-trained 2D neural networks for image feature map extraction. Denoting the image feature extraction function as $\mathcal{F}_I$, then image feature map $F_{I_k}$ for the $k$-th image are extracted and up-sampled to the original resolution as $I_k$ by,

$$F_{I_k} = \mathcal{F}_I(I_k) \quad (8)$$

where $F_{I_k} \in \mathbb{R}^{H\times W\times D_{2D}}$.

#### 3) 2D to 3D Alignment and Aggregation

Image feature maps are highly distinctive because pre-trained 2D neural networks contain informative knowledge from large-scale image datasets. This study further acquires point-wise 2D modality features from the $k$-th view $F_{2D,k} \in \mathbb{R}^{N\times D_{2D}}$ with 2D to 3D alignment according to $P_{2D,k}$,

$$F_{2D,k,i} = F_{I_k}(P_{2D,k,i}) \quad (9)$$

where $F_{2D,k,i}$ is the 2D modality feature for the $i$-th point in the $k$-th view, and $F_{I_k}(P_{2D,k,i})$ refers to the feature in pixel position $P_{2D,k,i}$ of $F_{I_k}$.

Then, point-wise features from different views $\{F_{2D,k}, k \in [1, N_V]\}$ are aggregated in a point-wise manner. In this study, an average pooling is utilized for aggregation,

$$F_{2D} = \frac{1}{N_V}\sum_{k=1}^{N_V} F_{2D,k} \quad (10)$$

where $F_{2D}$ is the aggregated point-wise 2D modality features composing cues from multiple views. The whole 2D modality PCD feature extraction process is depicted in Algorithm 1.

### E. Aggregation

For a given PCD $P_{3D}$, the extraction of 3D and 2D modality features have been depicted. To complement each other and further enhance the capability of features, this study normalizes and concentrates $F_{2D}$ and $F_{3D}$ together and acquire $F_{CPMF} \in \mathbb{R}^{N\times(D_{2D}+D_{3D})}$. Specifically, this study normalizes $F_{2D}$ and $F_{3D}$ into unit vectors first to ensure 3D and 2D modality features have the same magnitude and contribute similarly in later anomaly detection, which can be formulated as,

$$\widehat{F}_{2D,i} = \frac{F_{2D,i}}{\|F_{2D,i}\|_2}, \widehat{F}_{3D,i} = \frac{F_{3D,i}}{\|F_{3D,i}\|_2} \quad (11)$$

---

**Algorithm 1**: 2D modality feature extraction

**Require:** Input PCD: $P_{3D}$ ; Camera-related parameters: $C, K, T$ ; Number of views: $N_V$ ; PCD rotation matrixes $\{R_1, R_2, ..., R_{N_V}\}$.

**Ensure:** 2D modality point-wise features: $F_{2D}$

1: $\{P_{3D,k}, k \in [1, N_V]\}$ ←Rotate the input PCD $P_{3D}$ with $\{R_1, R_2, ..., R_{N_V}\}$ using (4).
2: $\{I_k, k \in [1, N_V]\}$ ←Render the rotated PCDs into multi-view images via $C, K, T$ using (5).
3: $\{P_{2D,k}, k \in [1, N_V]\}$ ← Calculate 2D-3D projection correspondences via $K, T$ using (6).
4: $\{F_{I_k}, k \in [1, N_V]\}$ ←Extract image feature maps with pre-trained 2D neural networks for $\{I_k, k \in [1, N_V]\}$ using (8).
5: $\{F_{2D,k}, k \in [1, N_V]\}$ ←Align the 2D feature maps to 3D space for point-wise feature extraction via $\{P_{2D,k}, k \in [1, N_V]\}$ using (9).
6: $F_{2D}$ ←Aggregate the multi-view 2D modality point-wise features $\{F_{2D,k}, k \in [1, N_V]\}$ using (10)
7: **return** $F_{2D}$

---

**Algorithm 2**: Anomaly detection using CPMF

**Require:** Training normal PCD set: $\mathbb{P}_{train}$ ; Unknown testing PCD set: $\mathbb{P}_{test}$;

**Ensure:** Anomaly scores list for $\mathbb{P}_{test}$: $\mathbb{S}$.

　　# Training Stage
1: $\mathbb{M}$ ←Initialize the memory bank.
2: **for** $P_{3D}$ in $\mathbb{P}_{train}$ **do**
3: 　$F_{3d}$ ←Extract 3D modality features using (1).
4: 　$F_{2D}$ ← Extract 2D modality features using Algorithm 1.
5: 　$F_{CPMF}$ ← Aggregate the 2D and 3D modality features using (11) and (12).
6: 　$\mathbb{M}$ ←Write $F_{CPMF}$ to the memory bank using (14)
7: **end for**
　　# Testing Stage
8: $\mathbb{S}$ ←Initialize the anomaly score list.
9: **for** $P_{3D}^{test}$ in $\mathbb{P}_{test}$ **do**
10: 　$F_{test}$ ←Extract CPMF features for $P_{3D}^{test}$ using (15)
11: 　$S$ ←Calculate point-wise anomaly scores using (16)
12: 　$\mathbb{S}$ ←Write $S$ to the anomaly scores set.
13: **end for**
14: **return** $\mathbb{S}$

---

where $\widehat{F}_{2D,i}$ and $\widehat{F}_{3D,i}$ are normalized features for the $i$-th point. Then, $\widehat{F}_{2D}$ and $\widehat{F}_{3D}$ are concentrated to get $F_{CPMF}$ as follows,

$$F_{CPMF} = \mathcal{C}(\widehat{F}_{2D}, \widehat{F}_{3D}) \quad (12)$$

where $\mathcal{C}(\cdot, \cdot)$ denotes to the concentration function.
In summary, denoting the function for extracting CPMF features as $\mathcal{F}_{CPMF}$, the extraction of $F_{CPMF}$ is simplified as

$$F_{CPMF} = \mathcal{F}_{CPMF}(P_{3D}) \quad (13)$$

and $F_{CPMF}$ is utilized to detect anomalies.

*F. PCD Anomaly Detection*

This study follows PatchCore [23] for anomaly detection. In the training phase, a memory bank $\mathbb{M}$ is constructed with all the $P_{3D} \in \mathbb{P}_{train}$,

$$\mathbb{M} = \bigcup_{P_{3D} \in \mathbb{P}_{train}} \mathcal{F}_{CPMF}(P_{3D}) \quad (14)$$

where $\mathbb{M}$ contains CPMF features for all points in the training PCD set.

In the testing phase, to infer anomaly scores for a testing PCD $P_{3D}^{test}$, this study extracts CPMF features for $P_{3D}^{test}$ and then calculate anomaly scores with the nearest neighbor search. The anomaly score calculation process can be formulated as follows,

$$F_{test} = \mathcal{F}_{CPMF}(P_{3D}^{test}) \quad (15)$$
$$S_i = \min_{f \in \mathbb{M}} \|F_{test,i} - f\|_2^2 \quad (16)$$

where $S \in \mathbb{R}^N$ is point-wise anomaly scores for $P_{3D}^{test}$, $S_i$ refers to the anomaly score for the $i$-th point. For the given testing PCD set $\mathbb{P}_{test}$, their corresponding anomaly scores list $\mathbb{S}$ can be acquired by (15) and (16).

The overall framework of CPMF and the corresponding anomaly detection process are summarized in Algorithm 2.

## IV. EXPERIMENTS

This section illustrates several sets of experiments that evaluate the performance of CPMF and demonstrate the influence of its components on anomaly detection.

*A. Experiment Settings*

　1) *Dataset*

This study conducts experiments on the MVTec 3D dataset [1], a recently published real-world multimodal anomaly detection dataset equipped with 2D RGB images and 3D PCD scans for ten categories. The dataset includes both deformable and rigid objects, partially with natural variations (*e.g.,* peach and carrot). While some defects can only be detected with RGB information, most defects in the MVTec 3D dataset are geometrical abnormalities. This study focuses on investigating PCD anomaly detection and only leverages 3D PCD scans in later experiments.

　2) *Implementation Details*

**Data Preprocessing:** For the point clouds in the MVTec3D dataset, this study follows BTF [3] to remove the nonsense background at first. Specifically, this study takes a ten-pixel wide strip around the image boundary to approximate the plane. After removing all NaNs (*i.e.,* noise) in the PCD, RANSAC [39] and DB-Scan [40] from the Open3D library [41] are applied to the strip for the plane approximation and used to filter the background plane.

**3D Modality Feature Extraction**: By default, this study follows BTF [3] and leverages FPFH [7] for 3D modality feature extraction. To speed up the calculations of FPFH, this study downsamples the PCD before calculating 3D modality features. The resulting 3D modality feature dimension $D_{3D}$ is 33.

TABLE I. QUANTITATIVE RESULTS (I-AUC) OF THE PROPOSED METHOD CPMF AND OTHER METHODS ON THE MVTEC3D DATASET. THE BEST IS IN BOLD, AND THE SECOND BEST IN UNDERLINED.

| Method | Bagel | Cable Gland | Carrot | Cookie | Dowel | Foam | Peach | Potato | Rope | Tire | Mean |
|---|---|---|---|---|---|---|---|---|---|---|---|
| Voxel GAN | 0.3830 | 0.6230 | 0.4740 | 0.6390 | 0.5640 | 0.4090 | 0.6170 | 0.4270 | 0.6630 | 0.5770 | 0.5376 |
| Voxel AE | 0.6930 | 0.4250 | 0.5150 | 0.7900 | 0.4940 | 0.5580 | 0.5370 | 0.4840 | 0.6390 | 0.5830 | 0.5718 |
| Voxel VM | 0.7500 | 0.7470 | 0.6130 | 0.7380 | 0.8230 | 0.6930 | 0.6790 | 0.6520 | 0.6090 | 0.6900 | 0.6994 |
| Depth GAN | 0.5300 | 0.3760 | 0.6070 | 0.6030 | 0.4970 | 0.4840 | 0.5950 | 0.4890 | 0.5360 | 0.5210 | 0.5238 |
| Depth AE | 0.4680 | 0.7310 | 0.4970 | 0.6730 | 0.5340 | 0.4170 | 0.4850 | 0.5490 | 0.5640 | 0.5460 | 0.5464 |
| Depth VM | 0.5100 | 0.5420 | 0.4690 | 0.5760 | 0.6090 | 0.6990 | 0.4500 | 0.4190 | 0.6680 | 0.5200 | 0.5462 |
| AST | 0.8810 | 0.5760 | 0.9560 | 0.9570 | 0.6790 | 0.7970 | 0.9900 | 0.9150 | 0.9560 | 0.6110 | 0.8318 |
| BTF (Depth iNet) | 0.6860 | 0.5320 | 0.7690 | 0.8530 | 0.8570 | 0.5110 | 0.5730 | 0.6200 | 0.7580 | 0.5900 | 0.6749 |
| BTF (Raw) | 0.6270 | 0.5060 | 0.5990 | 0.6540 | 0.5730 | 0.5310 | 0.5310 | 0.6110 | 0.4120 | 0.6780 | 0.5722 |
| BTF (HoG) | 0.4870 | 0.5880 | 0.6900 | 0.5460 | 0.6430 | 0.5930 | 0.5160 | 0.5840 | 0.5060 | 0.4290 | 0.5582 |
| BTF (SIFT) | 0.7110 | 0.6560 | 0.8920 | 0.7540 | 0.8280 | 0.6860 | 0.6220 | 0.7540 | 0.7670 | 0.5980 | 0.7268 |
| BTF (FPFH) | 0.8250 | 0.5510 | 0.9520 | 0.7970 | 0.8830 | 0.5820 | 0.7580 | 0.8890 | 0.9290 | 0.6530 | 0.7819 |
| **CPMF** | **0.9830** | **0.8894** | **0.9885** | **0.9910** | **0.9578** | **0.8094** | 0.9884 | **0.9590** | **0.9792** | **0.9692** | **0.9515** |

TABLE II. QUANTITATIVE RESULTS (P-PRO) OF THE PROPOSED METHOD CPMF AND OTHER METHODS ON THE MVTEC3D DATASET. THE BEST IS IN BOLD, AND THE SECOND BEST IN UNDERLINED.

| Method | Bagel | Cable Gland | Carrot | Cookie | Dowel | Foam | Peach | Potato | Rope | Tire | Mean |
|---|---|---|---|---|---|---|---|---|---|---|---|
| Voxel GAN | 0.4400 | 0.4530 | 0.8250 | 0.7550 | 0.7820 | 0.3780 | 0.3920 | 0.6390 | 0.7750 | 0.3890 | 0.5828 |
| Voxel AE | 0.2600 | 0.3410 | 0.5810 | 0.3510 | 0.5020 | 0.2340 | 0.3510 | 0.6580 | 0.0150 | 0.1850 | 0.3478 |
| Voxel VM | 0.4530 | 0.3430 | 0.5210 | 0.6970 | 0.6800 | 0.2840 | 0.3490 | 0.6340 | 0.6160 | 0.3460 | 0.4923 |
| Depth GAN | 0.1110 | 0.0720 | 0.2120 | 0.1740 | 0.1600 | 0.1280 | 0.0030 | 0.0420 | 0.4460 | 0.0750 | 0.1423 |
| Depth AE | 0.1470 | 0.0690 | 0.2930 | 0.2170 | 0.2070 | 0.1810 | 0.1640 | 0.0660 | 0.5450 | 0.1420 | 0.2031 |
| Depth VM | 0.2800 | 0.3740 | 0.2430 | 0.5260 | 0.4850 | 0.3140 | 0.1990 | 0.3880 | 0.5430 | 0.3850 | 0.3737 |
| 3D-ST128 | 0.9500 | 0.4830 | **0.9860** | **0.9210** | 0.9050 | 0.6320 | 0.9450 | **0.9880** | **0.9760** | 0.5420 | 0.8328 |
| BTF (Depth iNet) | 0.7690 | 0.6640 | 0.8870 | 0.8800 | 0.8640 | 0.2690 | 0.8760 | 0.8650 | 0.8520 | 0.6240 | 0.7550 |
| BTF (Raw) | 0.4010 | 0.3110 | 0.6380 | 0.4980 | 0.2500 | 0.2540 | 0.5270 | 0.5300 | 0.8080 | 0.2010 | 0.4418 |
| BTF (HoG) | 0.7110 | 0.7630 | 0.9310 | 0.4970 | 0.8330 | 0.5020 | 0.7430 | 0.9480 | 0.9160 | 0.8580 | 0.7702 |
| BTF (SIFT) | 0.9420 | 0.8420 | 0.9740 | 0.8960 | **0.9100** | 0.7230 | 0.9440 | 0.9810 | 0.9530 | 0.9290 | 0.9094 |
| BTF (FPFH) | **0.9730** | 0.8790 | 0.9820 | 0.9060 | 0.8920 | 0.7350 | 0.9770 | 0.9820 | 0.9560 | 0.9610 | 0.9243 |
| **CPMF** | 0.9576 | **0.9456** | 0.9793 | 0.8681 | 0.8974 | **0.7460** | **0.9795** | 0.9807 | 0.9610 | **0.9773** | **0.9293** |

*2D Modality Feature Extraction:* This study renders multi-view images for a given PCD using the implementations of the Open3D library [41], and the rendered images are fixed with a spatial resolution of $224 \times 224$. By default, this study leverages the first three blocks from ResNet18 [42] pre-trained on ImageNet [17] during 2D feature extraction, resulting in a 2D modality feature dimension $D_{2D}$ of 448. All images are normalized using the mean and variance of the ImageNet dataset before fed to the pre-trained networks. For the multi-view image rendering, the rotation matrixes are mainly determined by $\theta_x, \theta_y, \theta_z$, and $\theta_x, \theta_y, \theta_z$ are selected from a predefined small-angle view list $\{-\pi/16, 0, \pi/16\}$ to ensure visibility of most points. This study sequentially samples several images from the view list for 2D modality feature extraction with a default number of 27.

3) *Evaluation Metrics*

Image-level anomaly detection performance is measured via the area under the receiver-operator curve (AU-ROC) using produced anomaly scores. For simplicity, Image-level AU-ROC is denoted as I-ROC in this study. To measure anomaly segmentation performance, this study utilizes the Pixel-level PRO metric (denoted as P-PRO) [10], which considers the overlap of connected anomaly components. In line with prior work [2], [3], this study computes the per-class I-ROC and P-PRO values for comparisons.

B. *Comparisons with State-of-the-art Methods*

Table 1 and Table 2 summarize per-class comparisons between CPMF and other state-of-the-art methods, including baselines [1], AST [16], 3D-ST [43], and several benchmarking methods reported in BTF [3].

In terms of I-ROC, AST [16] achieves the best average performance in existing methods, but its performance (83.18% I-ROC) is still far from satisfactory. The proposed CPMF achieves significantly better performances (95.15% I-ROC) compared to existing methods. The performance of using $F_{CPMF}$ reaches the highest I-ROC in eight out of ten categories and the second highest I-ROC in the remaining two categories (*i.e.,* dowel and peach), effectively demonstrating the superiority of CPMF.

For the P-PRO criterion that reveals PCD anomaly localization performance, the previous work BTF has found that handcrafted descriptors were impressive in anomaly localization and achieved a significant P-PRO of 92.43% using FPFH features. CPMF outperforms BTF and obtains a P-PRO of 92.93%, demonstrating the strength of CPMF in localizing anomalies. Fig. 5 shows some selected qualitative anomaly detection results using CPMF, which clearly shows the effectiveness of CPMF in localizing geometrical abnormalities.

C. *Ablation Studies*

In this subsection, this paper shows the influence of individual components of CPMF, including the influence of the number of views for multi-view image rendering, the contribution of 2D and 3D modality features, and the influence of backbones. Fig. 6 compares the performances of CPMF with different numbers of views, different feature combinations, and different backbones. Notably, the performance of 3D modality features does not rely on the number of views or the backbones but is only determined by the leveraged 3D handcrafted descriptors, so it keeps the same in all situations.

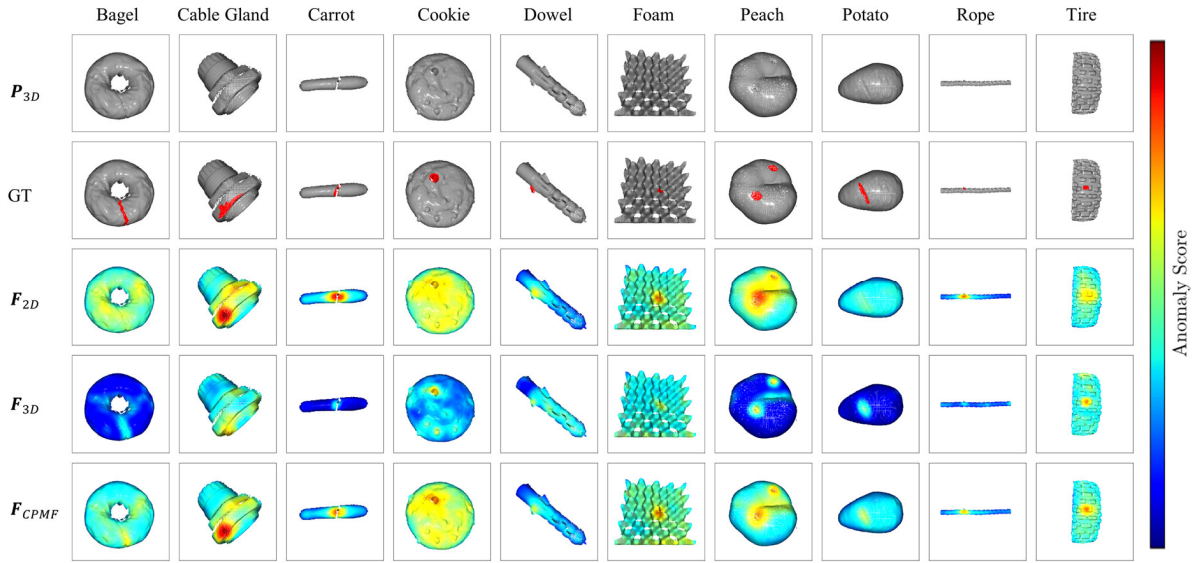

Fig. 5. Visualization of prediction results using different types of features ($F_{2D}, F_{3D}, F_{CPMF}(F_{2D}\&F_{3D})$). Best viewed in color.

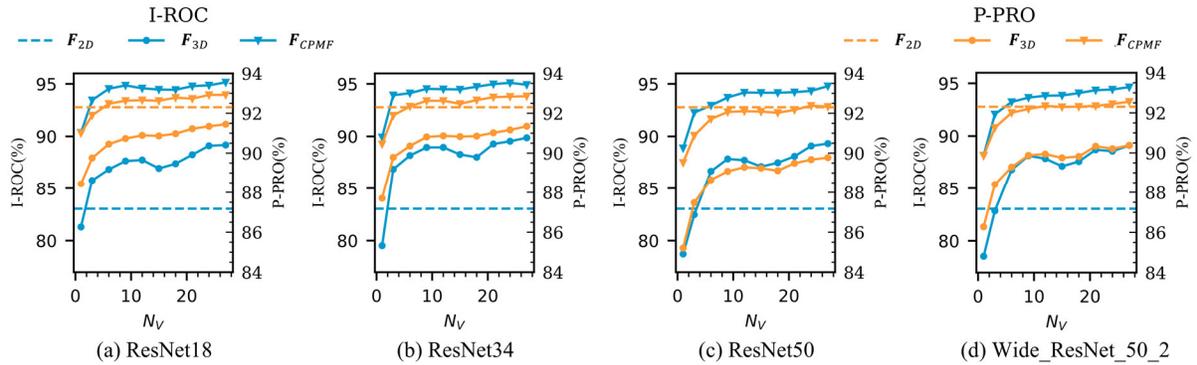

Fig. 6. Comparisons of the anomaly detection performances under different types of features ($F_{2D}, F_{3D}, F_{CPMF}(F_{2D}\&F_{3D})$), different number of views ($N_v$), and different backbones (ResNet18, ResNet34, ResNet50, Wide_ResNet_50_2).

1) *Influence of the number of rendering views*

Typically, a bigger $N_V$ indicates capturing information more comprehensively. To study the influence of $N_V$, this study conducts several sets of experiments using different $N_V$, $N_V \in \{1, 3, 6, \ldots, 27\}$. As Fig. 6 shows, I-ROC and P-PRO moderately improve with more views ($N_V$) and have the most dramatical rise when $N_V$ increases from one to three, with a slight drop when the rendering view number is within about 12 and 18.

For the overall improvements, it can be clearly explained that images from more views can better describe and capture the information underlying the given PCD, so better performances are achieved. Compared with $N_V = 1$, $N_V = 27$ brings significant improvements. For example, when using ResNet34 as the backbone of the pre-trained 2D neural networks, it sees improvements of about 10% I-ROC and 4% P-PRO using solely 2D modality features $F_{2D}$, and 5% I-ROC and 2% P-PRO using $F_{CPMF}$.

The main reason for the aforementioned slight drop may be that images from some specific views may result in low-quality

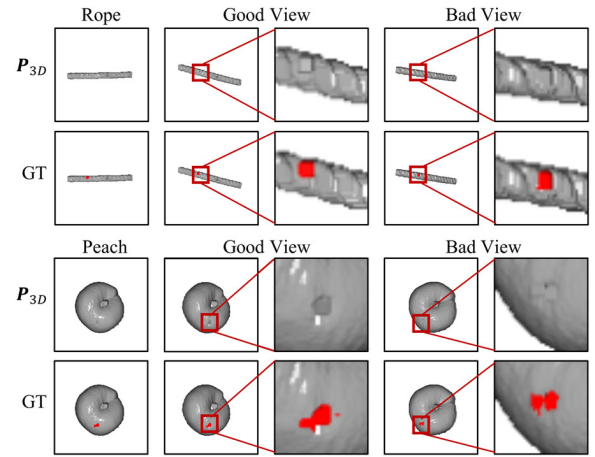

Fig. 7. Samples for the influence of different views. In certain good views, abnormalities are more visually obvious. By contrast, anomalies are hard to spot in the bad view. (Anomaly regions are highlighted in red)

features and damage anomaly detection performance. As many works [32], [34], [44] have demonstrated, adaptive views can better describe the structure of PCD, while fixed views may result in subpar performance. Therefore, selecting suitable views for 2D modality feature extraction could be further investigated to improve anomaly detection performance.

Fig. 7 shows two examples of images from different views. It effectively presents that abnormal regions are rendered visually differently in different views, and some views may provide better feature descriptiveness because of the apparent imaging of abnormalities in them. Although more views improve performance, Fig. 7 indicates that images from different views contribute differently to anomaly detection and could be selected adaptively for more substantial anomaly detection power.

2) *Influence of 2D and 3D modality features*

As previously discussed, 3D and 2D feature extraction modules describe PCD data in different ways and $F_{3D}$ and $F_{2D}$ contains different information, thus $F_{3D}$ composes abundant geometrical information and $F_{2D}$ exploits more semantics. Fig. 6 compares anomaly detection performances using different combinations of features. As can be seen, using solely $F_{3D}$ achieves a moderate average I-ROC (82.04%) and a great average P-PRO (92.30%) in all situations, while the figures for using solely $F_{2D}$ are significantly improved with more views $N_V$.

Concretely, in terms of I-ROC, using solely $F_{2D}$ underperforms using solely $F_{3D}$ when $N_V = 1$, but their combination $F_{CPMF}$ significantly performs better than using a single type of feature. With increasing of $N_V$, the performance of using solely $F_{2D}$ gradually improves and surpasses that of $F_{3D}$, indicating that multi-view images can better describe geometrical information in PCD. In addition, $F_{CPMF}$ continuously largely outperforms using a single type of feature (*e.g.*, with about a 6% improvement when using ResNet18), which effectively demonstrates the complementation between $F_{2D}$ and $F_{3D}$ again.

In terms of P-PRO, using solely $F_{2D}$ yields worse performance than using solely $F_{3D}$ in nearly all situations, with still a slight drop even using all views. The reason may be that $F_{2D}$ has a large receptive field than that of $F_{3D}$, resulting in weak geometrical point-wise features. $F_{CPMF}$ slightly exceeds using a single type feature, *e.g.*, by about 0.6% when using ResNet18. In summary, $F_{3D}$ performs better than $F_{2D}$ at the pixel level, while worse at the image level, proving that the extracted 3D modality features have stronger geometrical information and weaker semantics than 2D modality features. Their combination has both local geometrical and global semantic context and yields better image- and pixel-level performance.

Fig. 5 clearly shows that the $F_{2D}$ and $F_{3D}$ have different preferences in detecting PCD anomalies. For example, in the selected samples, solely $F_{2D}$ can well localize anomalies in categories like cable gland, carrot, dowel, etc., and performs weakly in categories including bagel and potato, whose anomalies can be well localized using pure $F_{3D}$ by contrast. The combination of $F_{2D}$ and $F_{3D}$, *i.e.*, the complete $F_{CPMF}$, significantly improves anomaly detection performance and localizes anomalies impressively in all categories, as Fig. 5 shows. Fig. 8 visualizes several feature distributions of $F_{2D}$, $F_{3D}$, and $F_{CPMF}$, and shows that a single type feature may not be distinguished well. By contrast, the distribution of $F_{CPMF}$ is more distinguishable.

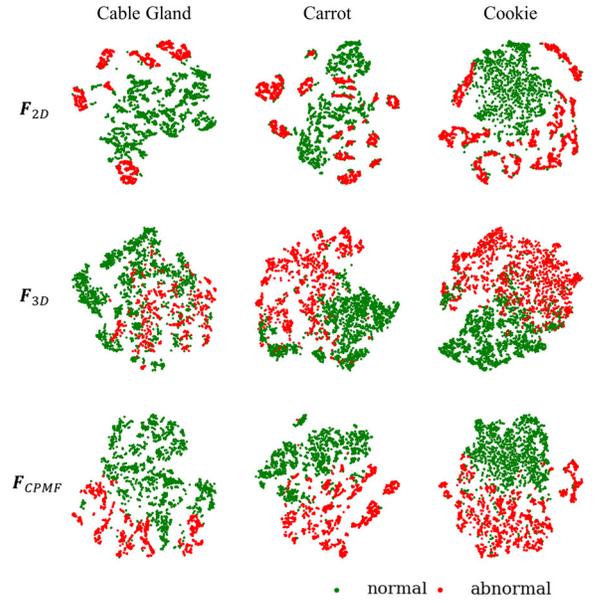

Fig. 8. Visualization for feature distributions under different feature combination.

3) *Influence of backbones*

Fig. 6 compares the performance of CPMF under different backbones, including ResNet18, ResNet34, ResNet50, and Wide_ResNet_50_2 [45], and Table 3 summarizes the best performances for different backbones. For various types of backbones, using solely $F_{2D}$ performs worse than using solely $F_{3D}$ in pixel-level while better in image-level, and the combination boosts both image- and pixel-level performances. In addition, the type of backbones does not greatly affect the performance, and CPMF achieves the best overall performance with ResNet18 (95.15% I-ROC and 92.93% P-PRO).

TABLE III. QUANTITATIVE RESULTS OF THE PROPOSED METHOD CPMF WITH DIFFERENT FEATURES AND BACKBONES.

| Backbone | Feature | I-ROC | P-PRO |
|---|---|---|---|
| \ | $F_{3D}$ | 0.8304 | 0.9230 |
| ResNet18 | $F_{2D}$ | 0.8918 | 0.9145 |
|  | $F_{CPMF}$ | 0.9515 | 0.9293 |
| ResNet34 | $F_{2D}$ | 0.8987 | 0.9135 |
|  | $F_{CPMF}$ | 0.9492 | 0.9286 |
| ResNet50 | $F_{2D}$ | 0.8932 | 0.8977 |
|  | $F_{CPMF}$ | 0.9479 | 0.9233 |
| Wide_ResNet_50_2 | $F_{2D}$ | 0.8911 | 0.9038 |
|  | $F_{CPMF}$ | 0.9464 | 0.9256 |

4) *Limitation analysis*

Several limitations are analyzed in this subsection. First, the quality of rendered images may be poor because of the noise injected during PCD acquisition, as is shown in Fig. 9. The poor quality may further influence the feature capability and lead to misjudgments. Second, the current pixel-wise criterion P-PRO reveals the performance of detecting all kinds of anomalies, but some anomalies can only be detected with RGB information. This phenomenon leads to the excellent but not saturated performance achieved by CPMF and appeals to a fairer metric for point-wise PCD anomaly localization.

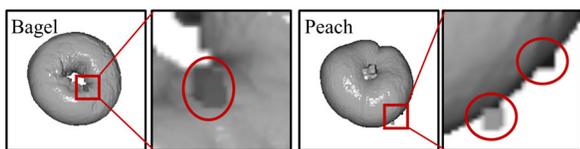

Fig. 9. Illustration for bad quality of rendered images resulting by acquisition noise.

## V. CONCLUSION

This paper addresses a promising and challenging task, *i.e.*, PCD anomaly detection, and proposes a novel PCD representation named Complementary Pseudo Multimodal Feature (CPMF) to achieve better anomaly detection performance. CPMF incorporates sophisticated handcrafted PCD descriptors and descriptive features from recently developed pre-trained 2D neural networks for thoroughly describing PCD data. Specifically, handcrafted PCD descriptors are directly applied to 3D modality to extract local geometrical information. To take advantage of pre-trained 2D neural networks, CPMF develops a 2D modality feature extraction module. The module renders PCD data into multi-view images via 3D to 2D projection and rendering. Then it employs pre-trained 2D neural networks to extract image feature maps, which are then aligned and aggregated to get the final 2D modality feature containing rich global semantics. The CPMF features composing 3D and 2D modality features exploit both local geometrical and global semantic cues, providing better descriptiveness. Comprehensive experiments demonstrate that the proposed CPMF outperforms previous methods by a large margin and proves the complementary capability between 2D and 3D modality features.

Despite the state-of-the-art performances CPMF achieved, there are still several promising future directions for exploration. For example, developing CPMF end-to-end can improve efficiency and reduce memory consumption. In addition, adaptive view selection and aggregation are expected to improve anomaly detection performance further.

## REFERENCES

[1] P. Bergmann, X. Jin, D. Sattlegger, and C. Steger, "The MVTec 3D-AD Dataset for Unsupervised 3D Anomaly Detection and Localization," in *Proceedings of the 17th International Joint Conference on Computer Vision, Imaging and Computer Graphics Theory and Applications*, 2022, pp. 202–213. doi: 10.5220/0010865000003124.
[2] P. Bergmann and D. Sattlegger, "Anomaly Detection in 3D Point Clouds using Deep Geometric Descriptors." arXiv, Feb. 23, 2022. Accessed: Jan. 12, 2023. [Online]. Available: http://arxiv.org/abs/2202.11660
[3] E. Horwitz and Y. Hoshen, "Back to the Feature: Classical 3D Features are (Almost) All You Need for 3D Anomaly Detection." arXiv, Nov. 28, 2022. Accessed: Nov. 29, 2022. [Online]. Available: http://arxiv.org/abs/2203.05550
[4] Y. Guo *et al.*, "Unsupervised Landmark Detection-Based Spatiotemporal Motion Estimation for 4-D Dynamic Medical Images," *IEEE Trans. Cybern.*, pp. 1–14, 2021, doi: 10.1109/TCYB.2021.3126817.
[5] Y. Cao, Q. Wan, W. Shen, and L. Gao, "Informative knowledge distillation for image anomaly segmentation," *Knowl.-Based Syst.*, vol. 248, p. 108846, Jul. 2022, doi: 10.1016/j.knosys.2022.108846.
[6] X. Ma *et al.*, "A Comprehensive Survey on Graph Anomaly Detection with Deep Learning," *IEEE Trans. Knowl. Data Eng.*, pp. 1–1, 2021, doi: 10.1109/TKDE.2021.3118815.
[7] R. B. Rusu, N. Blodow, and M. Beetz, "Fast Point Feature Histograms (FPFH) for 3D Registration," in *IEEE International Conference on Robotics and Automation*, 2009, pp. 3212–3217. doi: 10.1109/ROBOT.2009.5152473.
[8] T. Li, Q. Pan, L. Gao, and P. Li, "Differential evolution algorithm-based range image registration for free-form surface parts quality inspection," *Swarm Evol. Comput.*, vol. 36, pp. 106–123, Oct. 2017, doi: 10.1016/j.swevo.2017.04.006.
[9] C. Zhao, J. Yang, X. Xiong, A. Zhu, Z. Cao, and X. Li, "Rotation invariant point cloud analysis: Where local geometry meets global topology," *Pattern Recognit.*, vol. 127, p. 108626, Jul. 2022, doi: 10.1016/j.patcog.2022.108626.
[10] P. Bergmann, K. Batzner, M. Fauser, D. Sattlegger, and C. Steger, "The MVTec Anomaly Detection Dataset: A Comprehensive Real-World Dataset for Unsupervised Anomaly Detection," *Int. J. Comput. Vis.*, vol. 129, no. 4, pp. 1038–1059, Apr. 2021, doi: 10.1007/s11263-020-01400-4.
[11] J. Yang, Y. Shi, and Z. Qi, "Learning deep feature correspondence for unsupervised anomaly detection and segmentation," *Pattern Recognit.*, vol. 132, p. 108874, Dec. 2022, doi: 10.1016/j.patcog.2022.108874.
[12] V. Zavrtanik, M. Kristan, and D. Sko\vcaj, "Reconstruction by inpainting for visual anomaly detection," *Pattern Recognit*, vol. 112, p. 107706, 2020.
[13] Y. Shi, J. Yang, and Z. Qi, "Unsupervised anomaly segmentation via deep feature reconstruction," *Neurocomputing*, vol. 424, pp. 9–22, 2021, doi: https://doi.org/10.1016/j.neucom.2020.11.018.
[14] Q. Wan, L. Gao, X. Li, and L. Wen, "Unsupervised Image Anomaly Detection and Segmentation Based on Pre-Trained Feature Mapping," *IEEE Trans. Ind. Inform.*, pp. 1–10, 2022, doi: 10.1109/TII.2022.3182385.
[15] G. Wang, S. Han, E. Ding, and D. Huang, "Student-Teacher Feature Pyramid Matching for Anomaly Detection," in *British Machine Vision Conference*, Oct. 2021.
[16] M. Rudolph, T. Wehrbein, B. Rosenhahn, and B. Wandt, "Asymmetric Student-Teacher Networks for Industrial Anomaly Detection," in *Winter Conference on Applications of Computer Vision (WACV)*, Jan. 2023.
[17] O. Russakovsky *et al.*, "ImageNet Large Scale Visual Recognition Challenge," *Int. J. Comput. Vis.*, vol. 115, pp. 211–252, 2014.
[18] V. Zavrtanik, M. Kristan, and D. Sko\vcaj, "DRAEM – A discriminatively trained reconstruction embedding for surface anomaly detection," *2021 IEEECVF Int. Conf. Comput. Vis. ICCV*, pp. 8310–8319, 2021.
[19] P. Bergmann, S. Löwe, M. Fauser, D. Sattlegger, and C. Steger, "Improving Unsupervised Defect Segmentation by Applying Structural Similarity to Autoencoders," in *VISIGRAPP*, 2018, pp. 372–380.
[20] C.-L. Li, K. Sohn, J. Yoon, and T. Pfister, "CutPaste: Self-Supervised Learning for Anomaly Detection and Localization," *2021 IEEECVF Conf. Comput. Vis. Pattern Recognit. CVPR*, pp. 9659–9669, 2021.
[21] S. Niu, B. Li, X. Wang, S. He, and Y. Peng, "Defect attention template generation cycleGAN for weakly supervised surface defect segmentation," *Pattern Recognit.*, vol. 123, p. 108396, Mar. 2022, doi: 10.1016/j.patcog.2021.108396.
[22] D. Gudovskiy, S. Ishizaka, and K. Kozuka, "CFLOW-AD: Real-Time Unsupervised Anomaly Detection with Localization via Conditional Normalizing Flows," in *2022 IEEE/CVF Winter Conference on Applications of Computer Vision (WACV)*, Waikoloa, HI, USA, Jan. 2022, pp. 1819–1828. doi: 10.1109/WACV51458.2022.00188.
[23] K. Roth, L. Pemula, J. Zepeda, B. Scholkopf, T. Brox, and P. Gehler, "Towards Total Recall in Industrial Anomaly Detection," in *2022 IEEE/CVF Conference on Computer Vision and Pattern Recognition (CVPR)*, New Orleans, LA, USA, Jun. 2022, pp. 14298–14308. doi: 10.1109/CVPR52688.2022.01392.
[24] C. Steger, M. Ulrich, and C. Wiedemann, *Machine Vision Algorithms and Applications*. 2008.
[25] M. Bengs, F. Behrendt, J. Krüger, R. Opfer, and A. Schlaefer, "Three-dimensional deep learning with spatial erasing for unsupervised anomaly segmentation in brain MRI," *Int. J. Comput. Assist. Radiol. Surg.*, vol. 16, no. 9, pp. 1413–1423, Sep. 2021, doi: 10.1007/s11548-021-02451-9.
[26] Y. Song, L. Gao, X. Li, and W. Shen, "A novel robotic grasp detection method based on region proposal networks," *Robot. Comput.-Integr. Manuf.*, vol. 65, p. 101963, 2020, doi: https://doi.org/10.1016/j.rcim.2020.101963.
[27] A. Garcia-Garcia, F. Gomez-Donoso, J. Garcia-Rodriguez, S. Orts-Escolano, M. Cazorla, and J. Azorin-Lopez, "PointNet: A 3D Convolutional Neural Network for real-time object class recognition," in *2016 International*

*Joint Conference on Neural Networks (IJCNN)*, Vancouver, BC, Canada, Jul. 2016, pp. 1578–1584. doi: 10.1109/IJCNN.2016.7727386.
[28] C. R. Qi, L. Yi, H. Su, and L. J. Guibas, "PointNet++: Deep hierarchical feature learning on point sets in a metric space," *Adv. Neural Inf. Process. Syst.*, vol. 2017-Decem, pp. 5100–5109, 2017.
[29] L. Frittoli, D. Carrera, B. Rossi, P. Fragneto, and G. Boracchi, "Deep open-set recognition for silicon wafer production monitoring," *Pattern Recognit.*, vol. 124, p. 108488, Apr. 2022, doi: 10.1016/j.patcog.2021.108488.
[30] C. Choy, J. Park, and V. Koltun, "Fully Convolutional Geometric Features".
[31] S. Xie, J. Gu, D. Guo, C. R. Qi, L. Guibas, and O. Litany, "PointContrast: Unsupervised Pre-training for 3D Point Cloud Understanding," in *Computer Vision – ECCV 2020*, Cham, 2020, pp. 574–591.
[32] H. Su, S. Maji, E. Kalogerakis, and E. Learned-Miller, "Multi-view Convolutional Neural Networks for 3D Shape Recognition," in *2015 IEEE International Conference on Computer Vision (ICCV)*, Santiago, Chile, Dec. 2015, pp. 945–953. doi: 10.1109/ICCV.2015.114.
[33] H. Huang, E. Kalogerakis, S. Chaudhuri, D. Ceylan, V. G. Kim, and E. Yumer, "Learning Local Shape Descriptors from Part Correspondences with Multiview Convolutional Networks," *ACM Trans. Graph.*, vol. 37, no. 1, pp. 1–14, Jan. 2018, doi: 10.1145/3137609.
[34] Z. Han *et al.*, "3D2SeqViews: Aggregating Sequential Views for 3D Global Feature Learning by CNN With Hierarchical Attention Aggregation," *IEEE Trans. Image Process.*, vol. 28, no. 8, pp. 3986–3999, Aug. 2019, doi: 10.1109/TIP.2019.2904460.
[35] Y. Xu, C. Zheng, R. Xu, Y. Quan, and H. Ling, "Multi-View 3D Shape Recognition via Correspondence-Aware Deep Learning," *IEEE Trans. Image Process.*, vol. 30, pp. 5299–5312, 2021, doi: 10.1109/TIP.2021.3082310.
[36] T. Li, L. Gao, P. Li, and Q. Pan, "An ensemble fruit fly optimization algorithm for solving range image registration to improve quality inspection of free-form surface parts," *Inf. Sci.*, vol. 367–368, pp. 953–974, Nov. 2016, doi: 10.1016/j.ins.2016.07.030.
[37] Q. Wan, L. Gao, X. Li, and L. Wen, "Industrial Image Anomaly Localization Based on Gaussian Clustering of Pretrained Feature," *IEEE Trans. Ind. Electron.*, vol. 69, no. 6, pp. 6182–6192, Jun. 2022, doi: 10.1109/TIE.2021.3094452.
[38] H. Deng and X. Li, "Anomaly Detection via Reverse Distillation from One-Class Embedding." arXiv, Mar. 22, 2022. Accessed: Nov. 29, 2022. [Online]. Available: http://arxiv.org/abs/2201.10703
[39] M. A. Fischler and R. C. Bolles, "Random sample consensus: a paradigm for model fitting with applications to image analysis and automated cartography," *Commun ACM*, vol. 24, pp. 381–395, 1981.
[40] M. Ester, H.-P. Kriegel, J. Sander, and X. Xu, "A Density-Based Algorithm for Discovering Clusters in Large Spatial Databases with Noise," in *Knowledge Discovery and Data Mining*, 1996.
[41] Q.-Y. Zhou, J. Park, and V. Koltun, "Open3D: A Modern Library for 3D Data Processing," *ArXiv*, vol. abs/1801.09847, 2018.
[42] K. He, X. Zhang, S. Ren, and J. Sun, "Deep Residual Learning for Image Recognition," in *2016 IEEE Conference on Computer Vision and Pattern Recognition (CVPR)*, 2016, pp. 770–778. doi: 10.1109/CVPR.2016.90.
[43] P. Bergmann and D. Sattlegger, "Anomaly Detection in 3D Point Clouds using Deep Geometric Descriptors." arXiv, Feb. 23, 2022. Accessed: Nov. 23, 2022. [Online]. Available: http://arxiv.org/abs/2202.11660
[44] X. Wei, R. Yu, and J. Sun, "View-GCN: View-Based Graph Convolutional Network for 3D Shape Analysis," in *2020 IEEE/CVF Conference on Computer Vision and Pattern Recognition (CVPR)*, Seattle, WA, USA, Jun. 2020, pp. 1847–1856. doi: 10.1109/CVPR42600.2020.00192.
[45] S. Zagoruyko and N. Komodakis, "Wide Residual Networks," presented at the Proceedings of the British Machine Vision Conference, 2016, vol. abs/1605.07146, p. 87.1-87.12. doi: 10.5244/C.30.87.

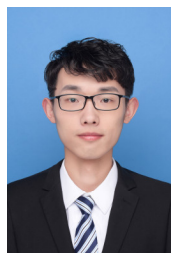

**Yunkang Cao** (Student Member, IEEE) was born in Jiangxi, China, in 1999. He received the B.S. degree from Huazhong University of Science and Technology (HUST), Wuhan, China, in 2020, where he is currently pursuing the Ph.D. degree in mechanical engineering.

His current research intesests include deep learning, anomaly detection and related applications in real industrial scenarios.

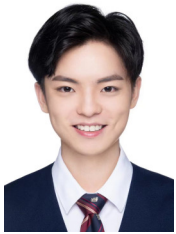

**Xiaohao Xu** received his B.S. degree in mechanical design, manufacturing and automation from Huazhong University of Science and Technology (HUST), Wuhan, China in 2022.

His current research interests include the fundamental theory and real-world applications of robotics, computer vision, and video understanding.

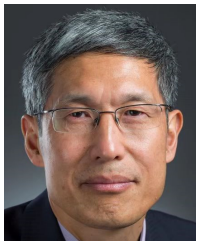

**Weiming Shen** (Fellow, IEEE) received the B.E. and M.S. degrees in mechanical engineering from Northern Jiaotong University, Beijing, China, in 1983 and 1986, respectively, and the Ph.D. degree in system control from the University of Technology of Compiegne, Compiegne, France, in 1996. He is currently a Professor with the Huazhong University of Science and Technology (HUST), Wuhan, China, and an Adjunct Professor with the University of Western Ontario, London, ON, Canada. Before joining HUST in 2019, he was a Principal Research Officer at the National Research Council Canada. He is a Fellow of Canadian Academy of Engineering and the Engineering Institute of Canada.

His work has been cited more than 16 000 times with an h-index of 61. He authored or coauthored several books and more than 560 articles in scientific journals and international conferences in related areas. His research interests include agent-based collaboration technologies and applications, collaborative intelligent manufacturing, the Internet of Things, and Big Data analytics